\documentclass[sigconf]{acmart} 

\AtBeginDocument{%
}

\copyrightyear{2026}
\acmYear{2026}
\setcopyright{cc}
\setcctype{by}
\acmConference[WWW '26]{Proceedings of the ACM Web Conference 2026}{April 13--17, 2026}{Dubai, United Arab Emirates}
\acmBooktitle{Proceedings of the ACM Web Conference 2026 (WWW '26), April 13--17, 2026, Dubai, United Arab Emirates}
\acmPrice{}
\acmDOI{10.1145/3774904.3792895}
\acmISBN{979-8-4007-2307-0/2026/04}

\usepackage{enumitem}
\usepackage{xspace}
\usepackage{subcaption}
\usepackage{multirow}
\usepackage{colortbl}
\usepackage{array}
\usepackage{booktabs}
\newcolumntype{C}[1]{>{\centering\arraybackslash}p{#1}}
\usepackage{xcolor}
\usepackage{soul}


\begin{document}

\title[ReFuGe: Feature Generation for Prediction Tasks on Relational Databases with LLM Agents]{ReFuGe:  Feature Generation for Prediction Tasks on Relational Databases with LLM Agents}

\settopmatter{authorsperrow=3}
        \author{Kyungho Kim} 
        \orcid{0009-0008-8304-9585}
	\affiliation{
    	\institution{KAIST}
            \city{Seoul}
            \country{Republic of Korea}
	}
	\email{kkyungho@kaist.ac.kr}

        \author{Geon Lee}
        \orcid{0000-0001-6339-9758}
	\affiliation{
		\institution{KAIST}
            \city{Seoul}
            \country{Republic of Korea}
	}
	\email{geonlee0325@kaist.ac.kr}

        \author{Juyeon Kim}
        \orcid{0009-0003-6102-6589}
	\affiliation{
		\institution{KAIST}
            \city{Seoul}
            \country{Republic of Korea}
	}
	\email{juyeonkim@kaist.ac.kr}

        \author{Dongwon Choi}
        \orcid{0009-0004-6693-6845}
	\affiliation{
		\institution{KAIST}
            \city{Seoul}
            \country{Republic of Korea}
	}
	\email{cookie000215@kaist.ac.kr}

        \author{Shinhwan Kang} 
        \orcid{0000-0001-6434-1347}
	\affiliation{
		\institution{KAIST}
            \city{Seoul}
            \country{Republic of Korea}
	}
	\email{shinhwan.kang@kaist.ac.kr}
	
	\author{Kijung Shin}
        \orcid{0000-0002-2872-1526}
	\affiliation{
		\institution{KAIST}
            \city{Seoul}
            \country{Republic of Korea}
	}
	\email{kijungs@kaist.ac.kr}

\begin{abstract}
Relational databases (RDBs) play a crucial role in many 
real-world web applications, supporting data management across multiple interconnected tables.
Beyond typical retrieval-oriented tasks, prediction tasks on RDBs have recently gained attention. 
In this work, we address this problem by generating informative relational features that enhance predictive performance. 
However, generating such features is challenging: it requires reasoning over complex schemas and exploring a combinatorially large feature space, all without explicit supervision.
To address these challenges, we propose \method, an agentic framework that leverages specialized large language model agents:
(1) a \textit{schema selection agent} identifies the tables and columns relevant to the task, (2) a \textit{feature generation agent} produces diverse candidate features from the selected schema, and (3) a \textit{feature filtering agent} evaluates and retains promising features through reasoning-based and validation-based filtering.
It operates within an iterative feedback loop until performance converges.
Experiments on RDB benchmarks demonstrate that \method substantially improves performance on various RDB prediction tasks. Our code and datasets are available at \url{https://github.com/K-Kyungho/REFUGE}.

\end{abstract}

\newcommand{\smallsection}[1]{\vspace{0.2pt}{\noindent {\bf{\underline{\smash{#1}}}}}}
\newtheorem{obs}{\textbf{Observation}}
\newtheorem{prp}{\textbf{Property}}
\newtheorem{dfn}{\textbf{Definition}}
\newtheorem{trm}{\textbf{Theorem}}

\newcommand\red[1]{\textcolor{red}{#1}}
\newcommand\blue[1]{\textcolor{blue}{#1}}
\newcommand\orange[1]{\textcolor{orange}{#1}}
\newcommand\brown[1]{\textcolor{brown}{#1}}
\newcommand\olive[1]{\textcolor{olive}{#1}}
\newcommand\sunwoo[1]{\textcolor{sunwooblue}{#1}}

\definecolor{mygreen}{rgb}{0,0.7,0}
\definecolor{sunwooblue}{RGB}{66, 133, 244}
\newcommand\green[1]{\textcolor{mygreen}{#1}}

\newcommand{\method}{\textsc{ReFuGe}\xspace}

\definecolor{verylightgray}{gray}{0.9}

\newcommand{\appropto}{\mathrel{\vcenter{
  \offinterlineskip\halign{\hfil$##$\cr
    \propto\cr\noalign{\kern1pt}\sim\cr\noalign{\kern-1pt}}}}}

\setlength{\textfloatsep}{0.10cm}
\setlength{\dbltextfloatsep}{0.10cm}
\setlength{\abovecaptionskip}{0.10cm}
\setlength{\skip\footins}{0.10cm}

\newcommand{\relbench}{\textsc{RelBench}\xspace}

\newcommand{\fone}{\texttt{rel-f1}\xspace}
\newcommand{\driverDNF}{\texttt{driver-dnf}\xspace}
\newcommand{\driverTopThree}{\texttt{driver-top3}\xspace}

\newcommand{\event}{\texttt{rel-event}\xspace}
\newcommand{\userIgnore}{\texttt{user-ignore}\xspace}
\newcommand{\userRepeat}{\texttt{user-repeat}\xspace}

\newcommand{\avito}{\texttt{rel-avito}\xspace}
\newcommand{\userClicks}{\texttt{user-clicks}\xspace}
\newcommand{\userVisits}{\texttt{user-visits}\xspace}

\newcommand{\stackex}{\texttt{rel-stack}\xspace}
\newcommand{\userEngage}{\texttt{user-engage}\xspace}
\newcommand{\userBadge}{\texttt{user-badge}\xspace}

\newcommand{\trials}{\texttt{rel-trial}\xspace}
\newcommand{\studyOutcome}{\texttt{study-outcome}\xspace}

\newcommand{\hm}{\texttt{rel-hm}\xspace}
\newcommand{\userChurn}{\texttt{user-churn}\xspace}

\newcommand{\amazon}{\texttt{rel-amazon}\xspace}
\maketitle

\vspace{-1mm}
\section{Introduction \& Related Works}
\label{sec:intro}
\begin{figure}
    \centering 
        \includegraphics[width=0.91\linewidth]{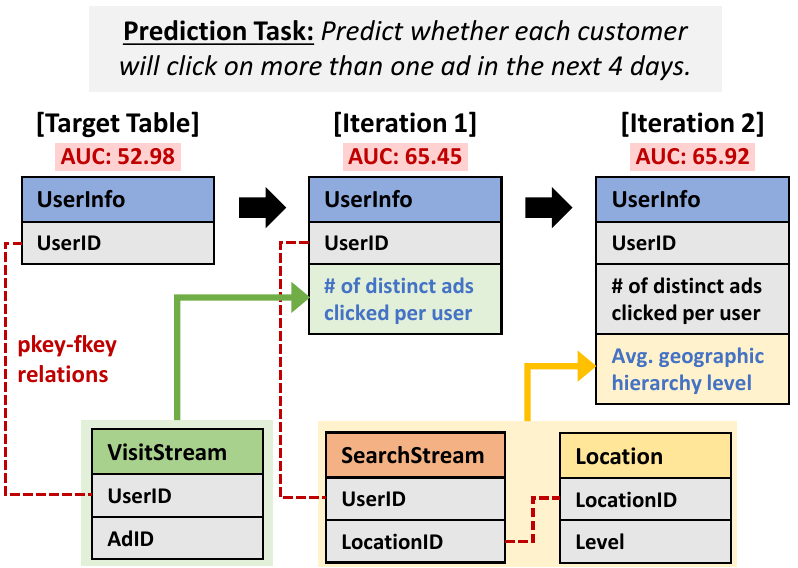}
        \caption{
        An example of feature generation by \method across iterations. 
        \method iteratively generates and appends relational features to the target table.
        }\label{fig:intro}
\end{figure}

\begin{figure*}[t]
    \centering
        \includegraphics[width=0.99\linewidth]{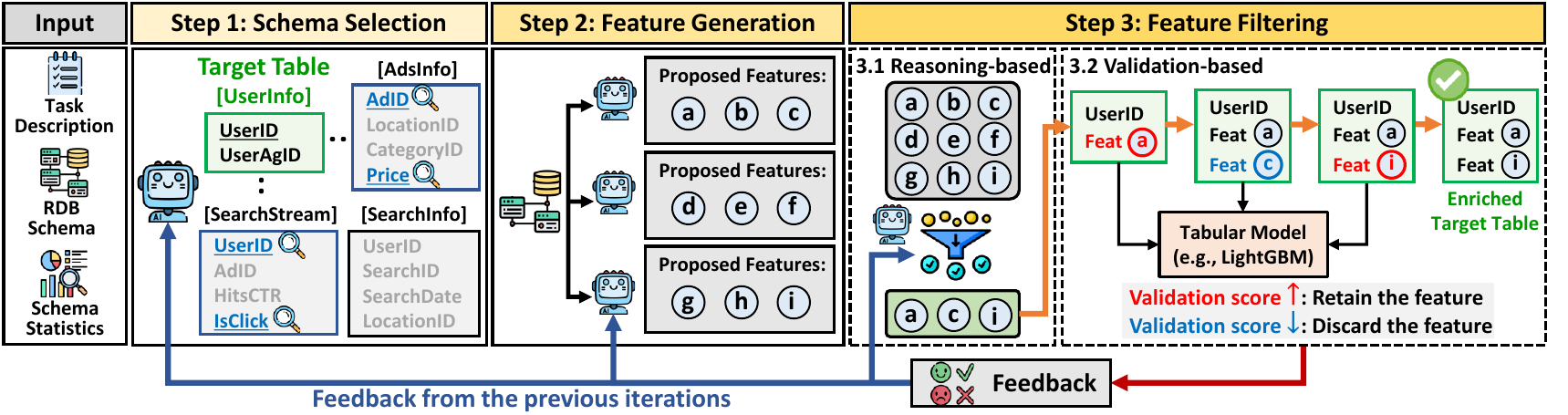}
        \caption{
        Overview of \method. 
        Through (1) schema selection, (2) feature generation, and (3) feature filtering, each performed by a specialized LLM agent. \method iteratively generates relational features that enrich the target table and improve performance on RDB prediction tasks. 
        At each iteration, feedback is generated and passed to the agents to guide their next rounds.
        \label{fig:model}}
\end{figure*}
Relational databases (RDBs) are widely used across various web applications, such as online retail platforms, social networking services, and forum platforms. 
Over the years, research on RDBs has made significant progress in \textit{retrieval-oriented} tasks, whose goal is to generate queries that extract relevant information~\cite{liu2024survey, katsogiannis2023survey, hong2024next}.
More recently, this scope has expanded to \textit{prediction} tasks (e.g., classification), aiming to exploit the relational structure of the RDB to improve downstream predictive performance~\cite{chen2025relgnn, fey2023relational, robinson2024relbench, wang20244dbinfer, dwivedi2025relational, choi2025rdb2g}.

A practical and effective approach for enabling prediction on RDBs is to generate features that capture predictive signals distributed across related tables~\cite{liang2024featnavigator}.
For example, a customer table can be enriched with aggregated purchase statistics, recent transaction counts, or attributes of associated accounts obtained from other tables. 
With this enriched table containing generated features, standard tabular learning methods~\cite{ke2017lightgbm, chen2016xgboost} can be applied directly.

However, automating effective feature generation in RDBs presents several challenges.
First, it requires understanding and extracting predictive signals from the relational structure \textbf{(C1)}. 
Second, the space of potential features is extremely large, due to the many ways in which tables can be transformed or combined \textbf{(C2)}.
Third, feature generation lacks ground-truth supervision, making direct supervised learning infeasible, unlike retrieval tasks (e.g., text-to-SQL) where ground-truth queries are available \textbf{(C3)}.

To address these challenges, we propose \textbf{\method} (\textbf{\underline{RE}}lational \textbf{\underline{F}}eat\textbf{\underline{U}}re \underline{\textbf{GE}}neration), 
an agentic framework that leverages the reasoning capabilities of LLMs.
To address (C1), we introduce a \textit{schema selection agent} that identifies the tables and columns in the RDB that provide the relational information most relevant to the prediction task.
For (C2), we employ multiple \textit{feature generation agents} that produce meaningful candidate features for the prediction table by leveraging domain knowledge and an understanding of the relational context.
The resulting candidate features are then selectively added to the table through reasoning-based and validation-based filtering steps.
Finally, to address (C3), we adopt an \textit{iterative self-improvement mechanism} that allows agents to refine schema selection and generate features without the ground-truth supervision. 
As shown in Figure~\ref{fig:intro}, \method progressively generates new features by leveraging information from multiple tables, leading to consistent improvements in validation AUROC over iterations.
Experiments on multiple RDBs demonstrate the effectiveness of \method in improving predictive performance.

Our key contributions are summarized as follows:
\begin{itemize}[leftmargin=*]
    \item \textbf{New Problem.} We introduce a new problem of generating features for prediction tasks on RDBs.
    \item \textbf{Agentic Framework.} We propose \method, an effective LLM-based agentic framework for feature generation in RDBs.
    \item \textbf{Empirical Validation.} We show that \method outperforms state-of-the-art competitors across seven datasets and eleven tasks.
\end{itemize}

\section{Task Formulation}
\label{sec:prelim}




We are given a relational database $\mathcal{R} = \{ \mathcal{T}_1, \mathcal{T}_2, \dots, \mathcal{T}_{|\mathcal{R}|} \}$, where each table $\mathcal{T}_i\in\mathcal{R}$ contains a set of entities (i.e., rows) and is connected to other tables through primary–foreign key relationships.
Informally, we assume that each table is a heterogeneous feature matrix $\mathcal{T}_i\in \mathcal{F}^{N_i\times d_i}$, where $N_i$ and $d_i$ denote the numbers of entities and features, respectively, and $\mathcal{F}$ denotes the space of admissible data types (e.g., numerical, categorical).

We consider a prediction task (e.g., classification) defined on the target entities of a designated target entity table $\mathcal{T}_\star\in\mathcal{R}$.
Our goal is to \textit{generate} additional features of $\mathcal{T}_\star$ by leveraging its relational context within the full relational database $\mathcal{R}$. The generated features are then concatenated with the original attributes of $\mathcal{T}_\star$ and used to train a tabular model (e.g., LightGBM~\cite{ke2017lightgbm}).

\vspace{-1mm}
\section{Proposed Framework}
\label{sec:method}
\renewcommand{\arraystretch}{1.0}
\begin{table*}[h]
\centering
\setlength{\tabcolsep}{2.6pt} 
\caption{(RQ1) Entity classification performance.
All metrics are reported in AUROC. The best results are highlighted in \textbf{bold}, and the second-best results are \underline{underlined}.}
\label{tab:performance}
\scalebox{0.94}{
\begin{tabular}{c|l|ccccccccccc|cc}
\toprule
 & \textbf{Data \& Task $\rightarrow$} & \multicolumn{2}{c}{\avito} & \multicolumn{2}{c}{\stackex} & \multicolumn{2}{c}{\fone} & \multicolumn{2}{c}{\event} & {\trials} & {\hm} & {\amazon} & \multicolumn{2}{c}{\textbf{Average}}\\
\cmidrule(lr){3-4} \cmidrule(lr){5-6} \cmidrule(lr){7-8} \cmidrule(lr){9-10} \cmidrule(lr){11-11} \cmidrule(lr){12-12} \cmidrule(lr){13-13} \cmidrule(lr){14-15}
 & \textbf{Method $\downarrow$} & {\texttt{visits}} & {\texttt{clicks}} & {\texttt{engage}} & {\texttt{badge}} & {\texttt{dnf}} & {\texttt{top3}} & {\texttt{repeat}} & {\texttt{ignore}} & {\texttt{outcome}} & {\texttt{churn}} & {\texttt{churn}} & {\textbf{AUROC}} & {\textbf{Rank}}\\
\midrule
\multirow{3}{*}{\centering \shortstack{\textbf{{Single}} \\ \textbf{{Table}}}} & {{LightGBM}}~\cite{ke2017lightgbm} & 53.05 & 53.60 & 63.39 & 63.43 & 68.56 & 73.92 & 68.04 & 79.93 & 70.09 & 55.21 & 52.22 & 63.77 & 6.8 \\
& {{XGBoost}}~\cite{chen2016xgboost} & 53.67 & 54.59 & 54.78 & 61.98 & 65.32 & 70.08 & 56.85 & 74.25 & 71.21 & 55.46 & 59.06 & 61.57 & 7.5\\
& {{FeatLLM}}~\cite{han2024large} & 51.15 & 48.69 & 42.56 & - & 70.43 & 52.72 & 46.67 & 47.88 & 57.35 & - & 55.11 & 52.51 & 8.7 \\
\midrule
\multirow{7}{*}{\centering \shortstack{\textbf{{RDB}}}} & {{ICL}}~\cite{wydmuch2024tackling} & 60.28 & 61.32 & 81.01 & 71.13 & 65.81 & \underline{88.47} & \underline{76.38} & 78.55 & 55.72 & 64.34 & 60.56 & 69.42 & 4.7  \\
& {{Rel-LLM}}~\cite{wu2025large} & 56.17 & \underline{62.28} & 69.46 & 62.12 & 71.84 & 70.64 & 68.12 & 61.32 & 59.02 & 55.95 & 60.07 & 63.36 & 6.0 \\
& {{RT-Gemma}}~\cite{ranjan2025relational}
& 62.70 & 59.80 & 78.00 & 80.00 & \underline{75.80} & \textbf{91.40} & - & - & 57.20 & 48.70 & 50.50 & 67.12 & 5.9 \\
& {{GPT}}~\cite{achiam2023gpt} & \underline{65.21} & 58.04 & \underline{83.82} & 80.21 & 65.16 & 77.04 & 71.26 & 76.41 & \textbf{72.94} & 55.50 & 63.72 & 69.94 & 4.3\\
& {{Claude}}~\cite{anthropic2025claude4} & 64.90 & 60.90 & 80.36 & 80.37 & 54.99 & 79.74 & 66.39 & 76.20 & 68.27 & 67.64 & \underline{64.66} & 69.49 & 4.7 \\
& {{LLM-CoT}}~\cite{kojima2022large} & 64.89 & 60.19 & 82.30 & \underline{80.97} & 61.41 & 75.92 & 70.13 & \underline{82.58} & 68.03 & \underline{68.06} & 64.42 & \underline{70.81} & \underline{4.2} \\
\cmidrule(lr){2-15}
 & \textbf{{\method (Ours)}} & \textbf{66.43} & \textbf{63.20} & \textbf{87.66} & \textbf{83.80} & \textbf{76.26} & 83.04 & \textbf{77.14} & \textbf{83.40} & \underline{72.22} & \textbf{68.10} & \textbf{67.00} & \textbf{75.30} & \textbf{1.3}  \\
\bottomrule
\end{tabular}}
\end{table*}

We propose \method, an agentic framework that generates relational features for prediction tasks on RDB (Figure~\ref{fig:model}). 
\method is composed of three specialized LLM agents: (1) a schema selection agent, (2) a feature generation agent, and (3) a feature filtering agent. 
These agents operate within an iterative feedback loop, progressively generating and refining features over iterations. 
\textbf{Detailed input prompts for every step are provided in Appendix~\cite{online2026appendix}.}

\smallsection{Step 1: Schema Selection.}
As an initial step, the \textit{schema selection agent} determines \textit{where} features should be generated from. 
The agent receives the RDB schema information, including tables, their column names, column types, and primary-foreign key relationships, along with the task description.
Using this information, it leverages its reasoning capabilities to identify the subset of tables and columns that are relevant to the prediction task, while filtering out those that are not. 
This step effectively reduces the search space from which features are generated in subsequent steps. 

\smallsection{Step 2: Feature Generation.}
Next, the \textit{feature generation agent} produces candidate features over the selected schema. 
Even after reducing the schema, the number of possible relational features remains exponentially large, with only a few of them likely to be useful.
To address this, the agent leverages its reasoning capabilities to propose features that are potentially informative and likely to improve predictive performance. 
To encourage diversity in the generated features, the agent employs multiple LLM instances,\footnote{We use the same model, with randomness ensured by a positive temperature.} each of which generates a potentially distinct set of candidate features.
These features are then aggregated into a unified feature pool. 



\smallsection{Step 3: Feature Filtering.}
Given the set of candidate features generated in Step 2, the \textit{feature filtering agent} applies a two-stage filtering process to retain only the most promising ones. 
\begin{itemize}[leftmargin=*]
    \item \textbf{Step 3.1. Reasoning-based Feature Filtering.} 
    The agent reviews the candidate features and, based on the task description and schema context, selects a smaller subset that is likely to be beneficial for the task. 
    Importantly, in the first iteration, the agent relies solely on its semantic reasoning over the provided information, whereas from the second iteration, it additionally considers the performance of previously generated features. 
    \item \textbf{Step 3.2. Validation-based Feature Filtering.} 
    Each selected feature is then (temporarily) added to the target table and used to train the tabular learning model (e.g., LightGBM).\footnote{In practice, we sample a subset of the training set rather than the full data for efficiency.} 
    If incorporating a feature improves validation performance, it is retained and appended to the target table; otherwise, it is discarded.
\end{itemize}

Notably, the two filtering stages complement each other. 
Reasoning-based filtering narrows the search space based on semantic reasoning, while validation-based filtering evaluates the empirical utility of each feature. 
Together, they ensure that the final features are both semantically meaningful and empirically effective. 


\smallsection{Iterative Feedback Loop.}
\method iteratively repeats Steps 1 - 3 until no additional features are selected. 
During this iterative loop, \method is designed to self-improve based on the outcomes of previous iterations. 
Specifically, at the end of the filtering step, it generates natural language feedback summarizing how each evaluated feature affected performance, e.g., ``\texttt{[Ad View Diversity]} increased validation AUROC from 65.06 to 66.30''. or ``{\texttt{[Direct Search Ratio]} caused validation AUROC to drop from 66.68 to 66.40}''. 
This feedback is then provided as input to each agent 
as a reference for guiding decisions in the subsequent iterations.
Feedback is accumulated across iterations, enabling the agents to self-learn which directions are promising and which should be avoided. 




\vspace{-1mm}
\section{Experiments}
\label{sec:experiments}

{\renewcommand{\arraystretch}{1.0}
\begin{table*}
\caption{(RQ2) Effectiveness of the key components of \method.
The best performance is highlighted in \textbf{bold}, and the second-best performance is \underline{underlined}.} \label{tab:ablation}
\setlength\tabcolsep{3.2pt}  
\scalebox{0.95}{
\begin{tabular}{l|ccccccccccc|cc}
    \toprule
    {\textbf{Data \& Task $\rightarrow$}} & \multicolumn{2}{c}{\avito} & \multicolumn{2}{c}{\stackex} & \multicolumn{2}{c}{\fone} & \multicolumn{2}{c}{\event} & {\trials} & {\hm} & {\amazon} & \multicolumn{2}{c}{\textbf{Average}} \\
    \cmidrule(lr){2-3} \cmidrule(lr){4-5} \cmidrule(lr){6-7} \cmidrule(lr){8-9} \cmidrule(lr){10-10} \cmidrule(lr){11-11} \cmidrule(lr){12-12}  \cmidrule(lr){13-14} 
    \textbf{Method $\downarrow$} & {\texttt{visits}} & {\texttt{clicks}} & {\texttt{engage}} & {\texttt{badge}} & {\texttt{dnf}} & {\texttt{top3}} & {\texttt{repeat}} & {\texttt{ignore}} & {\texttt{outcome}} & {\texttt{churn}} & {\texttt{churn}} & \textbf{AUROC}  & \textbf{Rank} \\
    \midrule
    \textbf{{\method-SS}} & {\textbf{66.61}} & 60.15 & 82.70 & {\textbf{83.89}} & {\underline{73.37}} & 79.46 
    & {\underline{73.34}} & {\textbf{88.36}} & {\underline{72.19}} & {\underline{68.04}} & 63.40 & \underline{73.77} & \underline{2.3}\\
    \textbf{{\method-FF}} & {\underline{66.56}} & 58.25 & {\textbf{88.51}} & 82.17 & 62.42 & 62.20 
    & 70.62 & 82.47 & 71.00 & {\underline{68.04}} & 63.94 & 70.56 & 3.0\\  
    \textbf{{\method-FB}} & 66.43 & {\underline{61.21}} & 87.61 & 63.41 & 62.40 
    & {\underline{81.32}} & 68.00 & 83.35 & 71.65 & 67.92 & {\underline{66.39}} & {70.88} & 3.1\\ 
    \midrule
    \textbf{{\method}} & 66.43 & {\textbf{63.20}} & {\underline{87.66}} & {\underline{83.80}} & {\textbf{76.26}} & {\textbf{83.04}} 
    & {\textbf{77.14}} & {\underline{83.40}} & {\textbf{72.22}} & {\textbf{68.10}} & {\textbf{67.00}} & {\textbf{75.30}} & \textbf{1.5}\\
    \bottomrule
\end{tabular}}
\vspace{2pt}
\end{table*}
}

In this section, we examine the following five research questions:
\begin{itemize}[leftmargin=*]
    \item \textbf{RQ1. Performance comparison.} How accurate is \method compared to existing baseline methods on RDB prediction tasks?
    \item \textbf{RQ2. Ablation study.} Do key components of \method contribute to its overall performance?
    \item \textbf{RQ3. Effect of iteration with feedback.} Does \method's performance improve through its feedback-driven iterative loop?
    \item \textbf{RQ4. Case study.} What features does \method generate?
    \item \textbf{RQ5. Parameter analysis.} How does the number of LLM instances in the feature generation agent impact performance?
\end{itemize}

\subsection{Experimental Settings}
We provide full details regarding the experimental settings in~\cite{online2026appendix}.

\smallsection{Datasets.}
We conduct experiments on seven real-world RDB benchmark datasets~\cite{robinson2024relbench}, spanning diverse domains. 

\smallsection{Baselines.}
We compare \method against both single-table and RDB methods.
\textit{Single-table} methods include 
(i) machine learning models~\cite{ke2017lightgbm, chen2016xgboost} and 
(ii) an LLM-based model~\cite{han2024large}.
These methods rely solely on the attributes of the target table and do not incorporate relational information.
\textit{RDB} methods include 
(iii) LLM-as-predictor approaches, where the LLM directly makes predictions using data joined from the RDB~\cite{wydmuch2024tackling, wu2025large, ranjan2025relational} and 
(iv) LLM-as-generator approaches, where a single LLM instance directly generates relational features using the same inputs as \method, but without schema selection, feature filtering, or a feedback loop~\cite{achiam2023gpt, anthropic2025claude4, kojima2022large}.
These methods leverage information from the RDBs. 


\subsection{Experimental Results}\label{sec:experiments:results}

\smallsection{(RQ1) Performance comparison.}
As shown in Table~\ref{tab:performance}, \method outperforms all baselines in 9 out of 11 tasks and achieves the best average performance and average rank overall. 
These results suggest that LLMs can effectively understand the relational structure of RDBs and generate informative features for prediction tasks. 
Notably, \method performs best not only on RDBs with simple schemas (e.g., \amazon and \hm with 3 tables and 2 primary–foreign key relations) but also on those with more complex schemas (e.g., \stackex with 7 tables with 12 primary–foreign key relations), demonstrating its applicability across varying schema complexities. 


\smallsection{(RQ2) Ablation study.}
To examine the effectiveness of the key components of \method, we consider three variants:
\begin{itemize}[leftmargin=*]
    \item \textbf{\method-SS}: Skips the schema selection step (step 2) and generates features directly from the full RDB schema.
    \item \textbf{\method-FF}: Replaces the reasoning-based feature filtering (step 3.1) with random selection of candidate features.
    \item \textbf{\method-FB}: Does not incorporate  feedback during iterations.
\end{itemize}
As shown in Table~\ref{tab:ablation}, \method outperforms all of its variants in 7 out of 11 cases, and achieves the best average rank.
Notably, \textbf{\method-FF} causes the largest performance drop, indicating the importance of reasoning over generated features and selecting the most promising ones.
\textbf{\method-FB} results in the second-largest drop, indicating that performance signals from earlier iterations provide meaningful guidance for subsequent feature generation.


\begin{figure}
    \centering
        \includegraphics[width=0.99\linewidth]{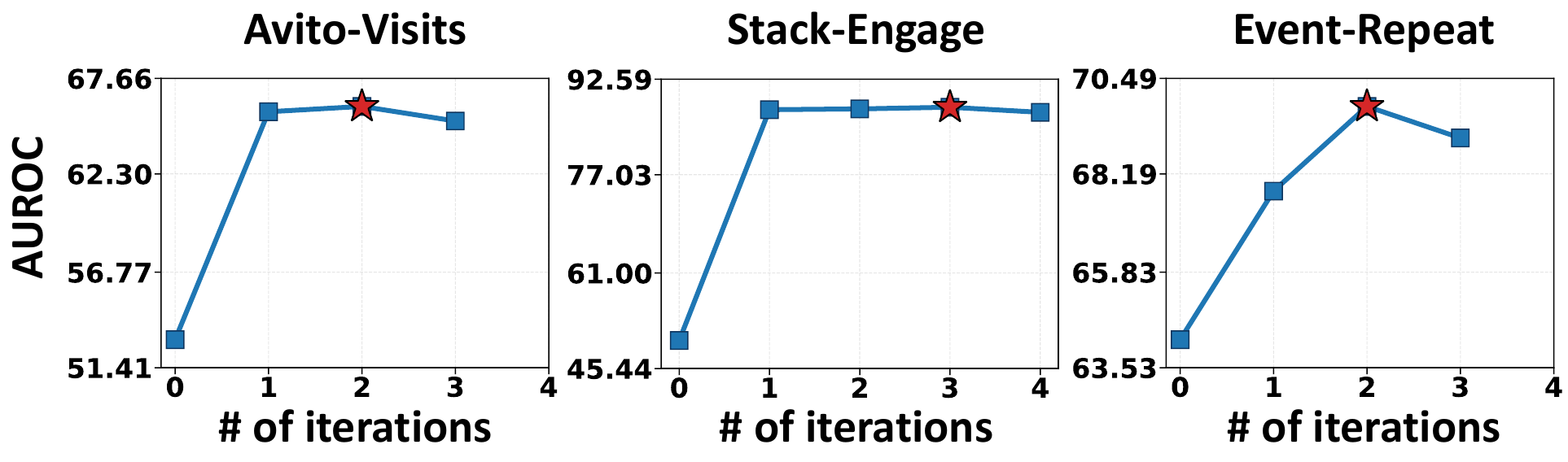}
        \caption{(RQ3) Evaluation performance tends to improve over iterations across all three tasks. The \red{$\star$} indicates the best validation performance over the iterations.} \label{fig:rq3}
\end{figure} 

\smallsection{(RQ3) Effect of iteration with feedback.}
As shown in Figure~\ref{fig:rq3}, the evaluation performance of \method generally improves over iterations, demonstrating the effectiveness of its feedback-driven iterative loop.
Iteration stops when no additional feature is selected, and \method performs an average of 2.4 iterations across all tasks.

\smallsection{(RQ4) Case study.}
In Figure~\ref{fig:intro}, we present a case study on the \avito dataset for the \userClicks task, which predicts whether each user in the target table 
will click on more than one advertisement in the next 4 days. 
\method first derives a feature capturing the number of distinct ads clicked per user by 
using the \textit{VisitStream} table, yielding a substantial performance gain. 
In the next iteration, it adds a feature capturing the average geographic hierarchy level of users’ search locations, obtained by joining the \textit{SearchStream} and \textit{Location} tables, which further improves performance. 

\begin{figure}[t]
    \centering
        \includegraphics[width=0.99\linewidth]{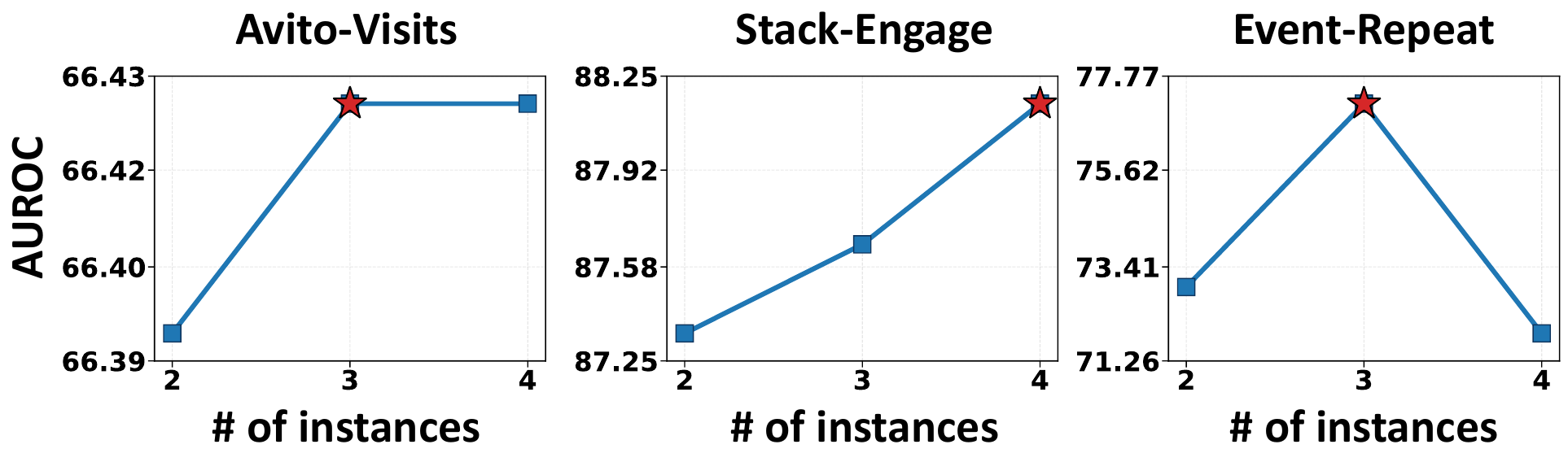}
        \caption{(RQ5) {Evaluation performance tends to increase as more LLM instances are used as feature-generation agents.} The \red{$\star$} indicates the best test performance.} \label{fig:rq5}
\end{figure}

\smallsection{(RQ5) Parameter analysis.}
As shown in Figure~\ref{fig:rq5}, we observe that using more LLM instances as the feature generation agent tends to improve performance across three tasks, with few exceptions. It indicates that increasing the diversity of generated candidate features is generally beneficial. In our experiments, we fix the number of instances to three for simplicity and efficiency. Additional parameter analyses are provided in~\cite{online2026appendix}.


\vspace{-1mm}
\section{Conclusions}
\label{sec:conclusion}

In this work, we address prediction on RDBs through the lens of feature generation.
We introduce a new problem of generating features for prediction tasks on RDBs (Section~\ref{sec:prelim}) and propose \method, an agentic framework that leverages LLM agents to iteratively generate and refine relational features based on feedback (Section~\ref{sec:method}). 
We conduct extensive experiments across multiple RDB benchmark datasets to demonstrate the effectiveness of \method (Section~\ref{sec:experiments}). 
For future work, we aim to extend our framework to additional downstream tasks, such as regression and link prediction.

\vspace{1mm} 
\smallsection{Acknowledgements.}
This work was supported by Institute of Information \& Communications Technology Planning \& Evaluation (IITP) grant funded by the Korea government (MSIT) (No. RS-2024-00438638, EntireDB2AI: Foundations and Software for Comprehensive Deep Representation Learning and Prediction on Entire Relational Databases, 60\%)
(No. RS-2022-II220157, Robust, Fair, Extensible Data-Centric Continual Learning, 30\%)
(No. RS-2019-II190075, Artificial Intelligence Graduate School Program (KAIST), 10\%).

\bibliographystyle{ACM-Reference-Format}
\bibliography{ref}

\end{document}